\documentclass[letterpaper, 10 pt, conference]{ieeeconf}  
\IEEEoverridecommandlockouts
\usepackage{graphicx}
\usepackage{cite}
\usepackage[export]{adjustbox}
\usepackage{amsmath} 
\usepackage[dvipsnames]{xcolor}
\usepackage{todonotes}
\usepackage{soul}

\usepackage[margin=1in]{geometry} 
\usepackage[colorlinks,citecolor=blue,urlcolor=blue,bookmarks=true,hypertexnames=true]{hyperref} 
\usepackage[hypcap=true,font=footnotesize,labelfont=bf]{caption}
\usepackage{lipsum}
\usepackage{float}
\makeatletter
\def\BState{\State\hskip-\ALG@thistlm}
\makeatother 
\usepackage[dvipsnames]{xcolor}
\usepackage{subfiles}
\usepackage{subcaption}
\usepackage{algpseudocode}
\usepackage{booktabs}
\usepackage[linesnumbered,ruled,vlined]{algorithm2e}

\newcommand{\acro}{SHY-COBRA}

\begin{document}
\title{Probabilistic Inference in Planning for Partially Observable Long Horizon Problems \\}
    
\author{Alphonsus Adu-Bredu$^{1}$ \hspace{0.5cm} Nikhil Devraj$^{1}$ \hspace{0.5cm}  Pin-Han Lin$^{1}$ \hspace{0.5cm} Zhen Zeng$^{2}$ \hspace{0.5cm} Odest Chadwicke Jenkins$^{1}$%

\thanks{$^{1}$Alphonsus Adu-Bredu, Nikhil Devraj, Pin-Han Lin and Odest Chadwicke Jenkins are with the Robotics Institute and Department of Electrical Engineering and Computer Science, University of Michigan, Ann Arbor, MI, USA.
        {\tt\small [adubredu|devrajn|pinhlin|ocj]@umich.edu}}
\thanks{$^{2}$Zhen Zeng is with J.P. Morgan AI Research. {\tt\small zhen.zeng@jpmchase.com} This work was completed independently from J.P. Morgan AI Research.
}}

\maketitle

\begin{abstract}
    For autonomous service robots to successfully perform long horizon tasks in the real world, they must act intelligently in partially observable environments. Most Task and Motion Planning approaches assume full observability of their state space, making them ineffective in stochastic and partially observable domains that reflect the uncertainties in the real world. We propose an online planning and execution approach for performing long horizon tasks in partially observable domains. Given the robot's belief and a plan skeleton composed of symbolic actions, our approach grounds each symbolic action by inferring continuous action parameters needed to execute the plan successfully. To achieve this, we formulate the problem of joint inference of action parameters as a Hybrid Constraint Satisfaction Problem (H-CSP) and solve the H-CSP using Belief Propagation. The robot executes the resulting parameterized actions, updates its belief of the world and replans when necessary. Our approach is able to efficiently solve partially observable tasks in a realistic kitchen simulation environment. Our approach outperformed an adaptation of the state-of-the-art method across our experiments.
     
\end{abstract}


\section{Introduction}
Autonomous service robots have the potential to perform long horizon tasks such as cooking meals in restaurants and homes and setting tables. In order for this potential to be realized, such robots would have to plan actions over large state and time horizons. They would also have to account for the uncertainties in their perception and knowledge of their environment. To ensure tractability, planning for such long horizon tasks is often decomposed into planning for symbolic actions and for continuous motions. The class of approaches that interleave symbolic and continuous planning is called integrated Task and Motion Planning (TAMP) \cite{hpn, ffrob, srivastava, garrettfactored, lgp, tampsurvey}. 

\begin{figure}
    \centering
    \includegraphics[width=\linewidth]{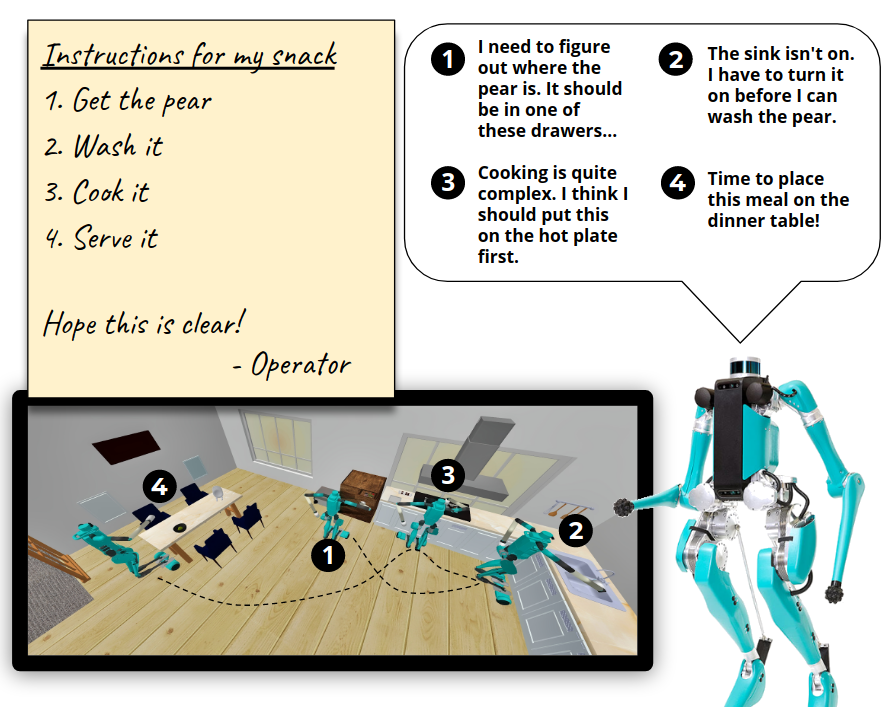}
    \caption{Given vague goals to make a snack, the robot generates and executes a coherent plan to successfully complete the assigned task.}
    \label{fig:teaser}
\end{figure} 


Major challenges that robots planning and acting in the real world face are the uncertainty in the robot's knowledge of the current state of the world and uncertainty in the effects of the robot's actions on the future state of the world. If these uncertainties are not accounted for when planning, the robot is likely to fail to accomplish the task at hand. Most Task and Motion Planning approaches \cite{hpn, ffrob, srivastava, garrettfactored, lgp, tampsurvey} assume full observability of their state space leading them to fail in  stochastic and partially observable domains that reflect the uncertainties in the real world.

We propose \emph{Satisfying HYbrid COnstraints with Belief pRopAgation} (\acro) as an approach for planning for long horizon partially observable TAMP problems. \acro{} takes as inputs the robot's noisy belief of the state of the world and a plan skeleton composed of symbolic actions that achieve a specified goal. \acro{} then infers satisfying parameter values for the actions needed to execute the plan successfully in partially observable domains. 

Given the robot's noisy belief of the state of the world and a plan skeleton composed of symbolic actions, \acro{}  formulates the problem of joint inference of action parameters as a Hybrid Constraint Satisfaction problem (H-CSP)\cite{tlpskeleton}, which is represented as a factor graph.  The factors in the factor graph are the action constraints whilst the variables are the symbolic and continuous action parameters to be inferred.  The continuous parameters are initialized by the robot's noisy belief of its environment. In most TAMP approaches  \cite{hpn, ffrob, srivastava, garrettfactored, lgp, tampsurvey}, the H-CSP is solved either by sampling or by constrained optimization. Neither of these approaches explicitly accounts for the uncertainty distributions of the continuous action parameters such as uncertainty in the pose estimates and in the robot's joint configurations. \acro{} instead solves the H-CSP using Pull Message Passing for Nonparametric Belief Propagation \cite{pmpnbp, karthikarticulated} because of its natural ability to account for the arbitrary uncertainty models of the continuous variables. The robot executes the actions in turn and replans when necessary.

We demonstrate our approach on several simulated partially observable long horizon tasks in a realistic simulation environment as shown in Figure \ref{fig:teaser}.
\section{Related Work} 
Our work focuses on the problem of planning for long horizon tasks. The class of approaches that interleave symbolic and continuous planning is called Task and Motion Planning (TAMP). Fully observable TAMP algorithms \cite{ffrob,hpn,lagriffoul2014efficiently,tlpskeleton,lgp, garrettfactored} assume that the agent has full knowledge of its deterministic domain and that the agent's actions have deterministic outcomes on its environment. These assumptions are however not representative of the kinds of domains robots operate in the real world, which are often stochastic and partially observable. In such domains, robots require the ability to plan in the face of incomplete knowledge and stochasticity in the effects of their actions.  Relatively few methods in TAMP literature have attempted to solve these types of TAMP problems \cite{lpk2013tampbelief}, \cite{abbeel2015modular}, \cite{phiquepal2019partial}, \cite{beliefpddlstream}.
 
Partially observable TAMP problems are often formulated as Hybrid Constraint Satisfaction Problems\cite{tlpskeleton} and solved either through sampling \cite{beliefpddlstream, abbeel2015modular} or constraint-optimization methods \cite{phiquepal2019partial}. Such approaches will often attempt to determinize the  belief via the Maximum Likelihood Observation \cite{abbeel2015modular, chadaxiomatic} rather than incorporating the entire distribution which provides richer information. Garrett et. al.(SS-Replan) \cite{beliefpddlstream} represent and update the belief over object poses using particle filtering. However, their approach limits the scope of partial observability to that of object poses. In contrast, through the use of Nonparametric Belief Propagation \cite{pmpnbp, karthikarticulated} on the constraint network, our approach provides the avenue for incorporating arbitrary uncertainty models on any of the variables whose value is to be inferred. We evaluate our approach against SS-Replan \cite{beliefpddlstream} in our experiments.

A number of prior works \cite{moon, decimation} have successfully solved Constraint Satisfaction Problems using Belief Propagation. Moon et. al. \cite{moon} formulates Sudoku Solving as a Constraint Satisfaction Problem and encodes it as a factor graph. The factors in this factor graph are row, column and 3x3 cell constraints and the variables are individual cells. They successfully infer satisfying cell values for empty cells in the sudoku puzzle. 

There exists much work in the domain of structuring planning problems as ones of inference \cite{toussaintstructured, BOTVINICK2012485, attias2003planning}. Toussaint et. al. \cite{toussaintstructured} uses structured Dynamic Bayesian Networks to represent structured planning domain and employs loopy belief propagation to solve short-horizon reaching  tasks under collision avoidance constraints with a humanoid upper body. Our approach formulates Task and Motion Planning as a Hybrid Constraint Satisfaction Problem (H-CSP) and uses Pull Message Passing Nonparametric Belief Propagation (PMPNBP) \cite{pmpnbp} to infer maximum joint beliefs that solve the H-CSP. 
\begin{figure}
\centering
     \includegraphics[width=0.47\textwidth]{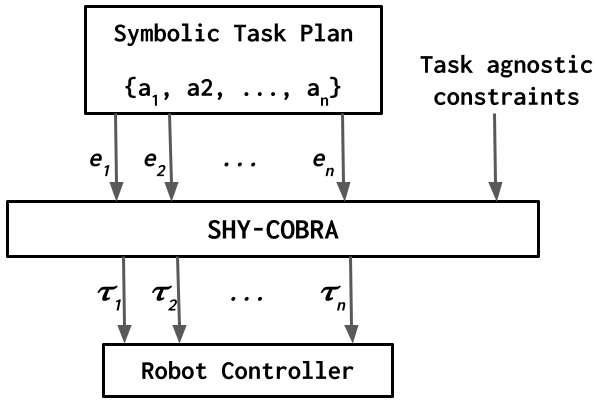}
      \caption{Outline of our approach. \{$e_1, e_2, \dots, e_n$\} are the symbolic effects of actions \{$a_1, a_2, \dots, a_n$\}. \{$\tau_1, \tau_2, \dots, \tau_n$\} are inferred trajectories sent to the robot controller for execution. }
       \label{fig:alg}
\end{figure}
\section{Problem Formulation}\label{sec:prob_form}

 Given an initial belief $\mathcal{B}_0=\{\mathcal{B}_0^{O}, \mathcal{B}_0^\phi\}$ of object poses and robot joint angles, and a symbolic plan skeleton $\{a_1, \cdots, a_n\}$, we aim to jointly ground each symbolic action $a_k$ into a robot pose $\phi_k$ in configuration space, along with the robot trajectory $\tau_{k, k+1}$ that takes the robot from pose $\phi_k$ to $\phi_{k+1}$. The robot can then sequentially execute the generated $\tau_{k,k+1}$ to achieve the end effect of each symbolic action $a_k$. After each trajectory execution, the robot perceives and updates the belief $\mathcal{B}_k$, and replans if the updated belief does not satisfy the desired end effect of the corresponding symbolic action.

We formulate the problem of jointly grounding a symbolic plan skeleton into a sequence of target robot poses $\{\phi_1, \cdots, \phi_n\}$ along with the in-between trajectories $\{\tau_{k,k+1} | k=[1, n-1]\}$ as a Hybrid Constraint Satisfaction Problem \cite{tlpskeleton}, where the constraints are imposed by the desired end effects of each symbolic action in the given plan, as well as other task-agnostic constraints such as collision-free and motion cost constraints.

\begin{figure}
\centering
     \includegraphics[width=0.45\textwidth]{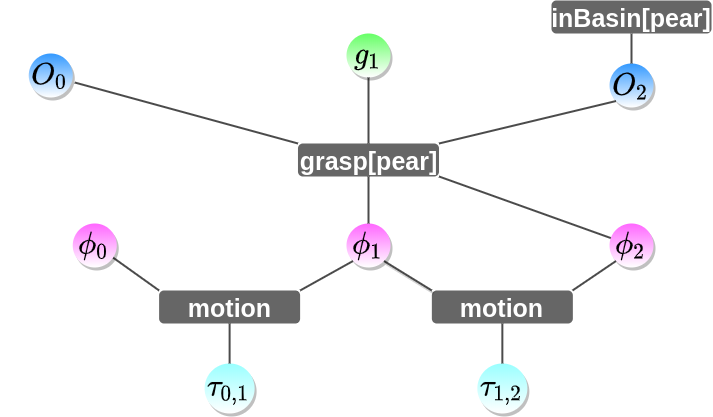}
      \caption{A Constraint Network. Given a task plan \{$a_1, a_2$\} composed of actions to pick up the pear ($a_1$), and to  transport the pear to the basin of a sink ($a_2$), we form a factor graph that imposes constraints (\textit{rectangular nodes}) as factors on the variable nodes (\textit{circular nodes}) that represent object poses, robot configurations and grasp poses. In this example, $\phi_0, \phi_1, \phi_2$ represent the initial robot configuration and target robot configuration for action $a_1, a_2$, respectively. $o_0, o_2$ represents the initial pear pose and target pear pose after the execution of action $a_2$. $g_1$ represents a grasp pose to grasp the pear. Note that $o_1$ is not included in this factor graph because it has the same value as $o_0$. Each target robot pose is also connected with factors that express the kinematic feasibility and collision-free constraints. We do not show these factors for clarity.  The grasp stability constraint on $g_1$ is also not shown in this diagram. The motion constraint node connected to variable nodes $\phi_0, \phi_1$ and $\tau_{(0,1)}$ encourages $\tau_{(0,1)}$ to be the shortest trajectory between configurations  $\phi_0$ and $\phi_1$.  PMPNBP is run on this factor graph to jointly infer satisfying action parameters.}
       \label{fig:const_network}
\end{figure}

\begin{algorithm}
\SetAlgoLined
\KwIn{High-level Goals, $G$, Initial Belief, $\mathcal{B}_0$}
\SetKwFunction{FMain}{\acro}
\SetKwProg{Fn}{Function}{:}{}
$\Pi ~\leftarrow $ \textsc{SymbolicPlanner} ($G$) \Comment{Generate a plan skeleton $\Pi$ that achieves goal $G$ using a symbolic planner} \\
\Fn{\FMain{$\Pi$, $\mathcal{B}_0$}}{  
    
    $G  ~\leftarrow$ \textsc{ConstraintNetwork}($\Pi$) \Comment{\textit{Convert $\Pi$ into Constraint Network $G$}}\\
    $G_{init}  ~\leftarrow$ \textsc{Initialize} ($G, \mathcal{B}_0$) \Comment{\textit{Initialize variable nodes in $G$ with $\mathcal{B}_0$}}\\
    $G_{conv}  ~\leftarrow$ \textsc{PMPNBP} ($G_{init}$) \Comment{\textit{Pass messages across $G_{init}$ using PMPNBP algorithm until convergence}}\\ 
    $\Pi_s ~\leftarrow$ \textsc{MaxSamples} ($G_{conv}, \Pi$) \Comment{\textit{Set variable params in $\Pi$ with the max-product assignment from corresponding nodes in $G_{conv}$}}\\ 
    \ForEach{$a \in \Pi_s$}
    {
        $o \leftarrow$ \textsc{ExecuteAction}($a$) \Comment{\textit{Receive observation $o$}}\\ 
        $\mathcal{B}_{current} \leftarrow$ \textsc{UpdateBelief}($a, o$) \\
        
         \If{$o \neq$ \textsc{ExpectedEffect}($a$) } { 
            
             $\Pi ~\leftarrow$ \textsc{UpdatePlanSkel.} ($\Pi_s, a$) \Comment{\textit{Update plan skeleton to reflect current state after executing $a$}}\\
             \textsc{\acro}($\Pi, \mathcal{B}_{current}$)\\
             $\textbf{return}$ \\
            } 
    }
\textbf{return} 
}
\textbf{End Function}
\caption{\acro}
\label{alg:bppai}
\end{algorithm}
 
\section{Methodology}

\subsection{Satisfying Hybrid Constraints with Belief Propagation (\acro)} 
As described in Algorithm \ref{alg:bppai} and Figure \ref{fig:alg},  \acro{} takes as input, the robot's noisy belief of the state of the world and a plan skeleton \cite{tlpskeleton} composed of symbolic actions. The plan skeleton is obtained by a symbolic planner that plans actions to achieve specified goal(s). Actions in this plan skeleton have free parameters like grasp poses and arm trajectories whose values are needed to be able to execute these actions in the world. We formulate the problem of inferring the values of the free parameters as a Hybrid Constraint Satisfaction Problem (H-CSP) (Section \ref{sec:hcsp}) and represent it as a constraint network as shown in Figure \ref{fig:const_network}. The variable nodes in the constraint network are initialized with uniformly weighted sets of samples that represent the robot's noisy belief of the value of the corresponding variable. For instance, a variable node representing a target object's pose is initialized by a set of uniformly weighted poses that represent the uncertainty distribution of the pose as estimated by the robot's noisy perception system. 

The  H-CSP is solved by inferring the satisfying maximum joint belief of the variable nodes using a max-product version of Pull Message Passing Nonparametric Belief Propagation (PMPNBP)\cite{pmpnbp} (Section \ref{sec:solve_hcsp}). We then assign each free parameter in the plan skeleton with the max-product assignment of the belief of its corresponding variable node and execute the plan in the world. 

If there is an unexpected effect of the robot's action while executing the plan, we update the plan skeleton and the constraint network and perform message passing using PMPNBP to infer the variable assignments for the new constraint network. This process continues until the robot successfully completes the task.  

The following subsections describe the components of \acro{} in detail.

\subsection{Hybrid Constraint Satisfaction Problem}\label{sec:hcsp}
Finding values for action parameters in a plan skeleton that satisfy all the sets of constraints is a \textit{Hybrid Constraint Satisfaction Problem} (H-CSP). The joint set of action parameters and constraints of a plan skeleton form a factor graph called a constraint network \cite{dechter1992constraint, lagriffoul2014efficiently}. A constraint network is a factor graph with constraints as Factor nodes and action parameters as Variable nodes. Edges exist between constraints and their corresponding action parameters as depicted in Figure \ref{fig:const_network}. 
Formally, an H-CSP is represented as
\begin{equation} \label{eq:fg}
G(X_1, X_2, ...,X_r) \sim \prod_{j=1}^N f_j (S_j)
\end{equation}
where $S_j \subseteq \{X_1, ...,X_r\}$, a subset of action parameter variable nodes that are subject to constraint factor node $f_j$, $G$ is the factor graph and $N$ is the number of factors. 

To solve the H-CSP is to infer values for all action parameters that satisfy their corresponding constraints. In this work, we propose to solve the H-CSP by performing max-product Nonparameteric Belief Propagation on the constraint network that represents the plan skeleton. The following subsections describe the Nonparametric Belief Propagation algorithm we use, our message-passing scheme and how the values of the action parameters are inferred from noisy observations and partial knowledge of the robot's environment.

\begin{figure*}
\centering
      \includegraphics[width=0.6\textwidth]{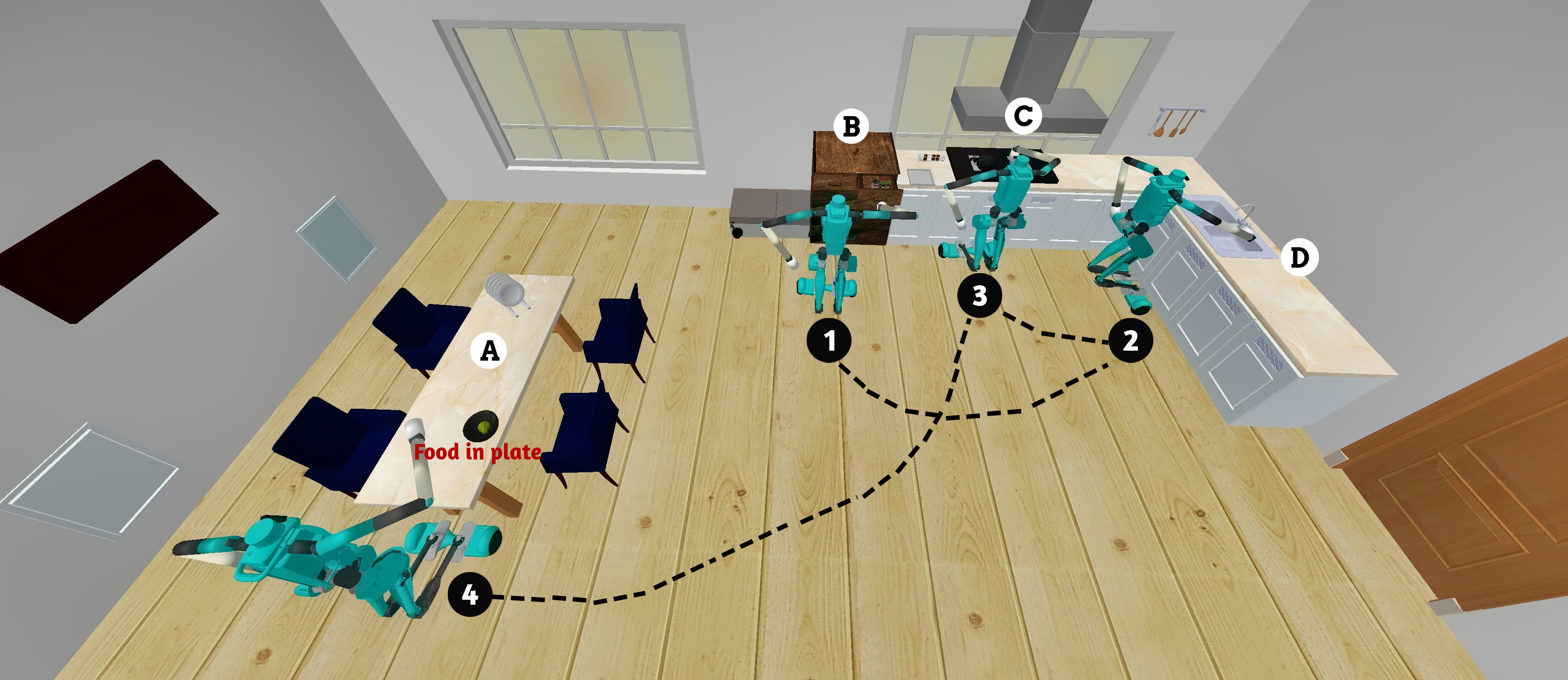}
      \caption{The kitchen simulation environment. The area annotated \textbf{(A)} is the dining area, \textbf{(B)} is the grocery cabinet, \textbf{(C)} is the cooking area and \textbf{(D)} is the sink area. To make a pear dinner according to an optimal plan from the \acro{} planner, the robot first moves to the grocery cabinet, opens one of the drawers in the cabinet and inspects it for the pear. If the pear is not found in the drawer, the robot updates its belief of the location of the pear and replans to inspect a different drawer. If the pear is found, the robot \textbf{(1)} picks up the pear, \textbf{(2)} moves to the sink, turns on the tap and washes the pear. The robot then \textbf{(3)} moves to the stove and cook the pear. After cooking the pear, the robot \textbf{(4)} finally moves to the dining table and serves the meal.}
       \label{fig:serve}
\end{figure*}

\subsection{Solving the H-CSP using PMPNBP}\label{sec:solve_hcsp}
To solve the H-CSP, we use PMPNBP to infer action parameter values that satisfy all the constraints in the constraint network. Mathematically, this is equivalent to inferring the action parameter assignments that maximize the joint probability 
\begin{equation} \label{eq:joint}
P(X) = \frac{1}{Z}\cdot \prod_{j=1}^N f_j (S_j), S_j \subseteq \{X_1, ...,X_r\}
\end{equation}
where $Z$ is a normalizing constant, $S_j \subseteq \{X_1, ...,X_r\}$, a subset of action parameter variable nodes that are subject to constraint factor node $f_j$ and $N$ is the number of factors. In the context of our work, $X$s include the robot pose $\phi_k$s in configuration space as well as robot trajectories $\tau_{k,k+1}$ as discussed in Section~\ref{sec:prob_form}.

To infer satisfying variable assignments, we perform message passing on the constraint network with max-product PMPNBP \cite{pmpnbp}. At the beginning, the belief of each variable node is initialized with uniformly weighted samples generated by specialized generators operating on the robot's initial belief. These generators used in this work are described in detail in Section \ref{sec:init_wtsamp}.

Message passing on a constraint network involves two kinds of messages; the \texttt{Constraint-to-variable} message and the \texttt{Variable-to-constraint} message. 

The \texttt{Constraint-to-variable} message $msg_{f(x) \rightarrow x}$ for iteration $m$ is computed as
\begin{equation*}
    \{\mu_x^{(i)}\}_{i=1}^M \sim bel^{m-1}(x) 
\end{equation*}
\begin{equation*}
    \{w_x^{(i)}\}_{i=1}^M  = \{ \text{max}_{\substack{y \in \rho(f) \setminus \{x\} \\ j \in 1, \cdots, M}}~ \sigma_f (\mu_x^{(i)}, \mu_y^{(j)})\}_{i=1}^M
\end{equation*}
\begin{equation}
    msg^m _{f \rightarrow x} = \{(\mu_{x}^{(i)}, w_{x}^{(i)}) : 1 \leq i \leq M\}
\end{equation}
where $y \in \rho(f) \setminus \{x\}$ represents the messages from variable nodes with edges to constraint node $f$ except variable node $x$

As formulated above, $M$ samples are drawn from the belief distribution of node $x$ from the previous iteration, $m-1$. The constraint function $\sigma_f$ of the constraint node $f$ is then used to compute weights for each of these samples. These weighted samples, which are now the \texttt{Constraint-to-variable} message are then passed to node $x$. Details on constraint functions can be found in Section \ref{sec:cnstfxn}.

The \texttt{Variable-to-constraint} message $msg_{x \rightarrow f(x)}$ for iteration $m$ is computed as 
\begin{equation}
    msg^m _{x \rightarrow f} = \bigcup _{y \in \rho(x) \setminus \{f\}} msg^{m-1} _{y \rightarrow x}
\end{equation}
Where $\rho(x) \setminus \{f\}$ are all the constraint nodes with edges to the variable node $x$ except constraint node $f$.

As formulated above, to approximate the product of incoming messages, we take the union of all incoming messages from neighboring constraint nodes except constraint node $f$ and normalize their weights. The resulting messages are then resampled and passed to constraint node $f$. 

One iteration of message passing on the constraint network follows the following sequence:
\begin{enumerate}
\item Pass \texttt{Variable-to-constraint} messages from all variables to their corresponding constraints
\item Pass \texttt{Constraint-to-variable} messages from all constraints to their corresponding variables
\end{enumerate}

After each iteration of message passing, the belief of each variable node is updated by taking the union of all incoming messages to the variable node, normalizing their weights and resampling a new set $\{\mu_x^{(i)}\}_{i=1}^M$ to represent the belief of the variable node.

Message passing is performed for several iterations until the maximum joint belief of each variable node converges.

After convergence, each action parameter $x_i$ is assigned with the max-product assignment of the belief of its variable node. 

\subsection{Constraints and Constraint functions}\label{sec:cnstfxn}
\subsubsection{Constraints}
The constraints used in this work are
\begin{itemize}
    \item \texttt{Motion} constraints that enforce that $\tau$ is the shortest trajectory from one robot configuration $\phi_0$ to another robot configuration $\phi_1$
    \item \texttt{Kin} constraints that enforce kinematic feasibility of the robot in configuration $\phi$ holding an object with grasp $g$
    \item \texttt{CfreeH} constraints that enforce that when the robot is holding an object in grasp $g$, the trajectory $\tau$ that the object is moved through is collision free
    \item \texttt{GraspH} constraints that enforce that a grasp $g$ is a stable grasp pose
    \item \texttt{grasp} constraints that enforce that at a configuration $\phi_0$, when a robot picks an object at position $p0$ with grasp $g0$, the grasp $g0$ will be feasible at a later time when the robot at configuration $\phi_1$ places the object at position $p1$.
    \item \texttt{Stable} constraints that enforce that a placement pose $p$ of an object is stable and won't cause the object to fall off.
    \item \texttt{inBasin} constraints that enforce that the object is placed in a stable pose in the basin
    \item \texttt{inSaucepan} constraints that enforce that the object is placed in a stable pose on the saucepan.
\end{itemize}

\subsubsection{Constraint functions}
A constraint function assigns weights to samples drawn from the belief distribution of the target variable node when the \texttt{Constraint-to-variable} message is computed.
The weight of each sample drawn from a variable node $a$ is computed as follows:
\begin{equation}
    w_a ^i = \sigma_f (x_a ^i, \hat{x}_1, \hat{x}_2, \dots, \hat{x}_{T-1})
\end{equation}
where $w_a ^i$ is the weight of sample $i$ drawn from variable node $a$, $\sigma_f$ is the constraint function of constraint node $f$, $T$ is the number of variable nodes connected by an edge to the constraint node $f$ and $\hat{x}_1, \hat{x}_2, \dots, \hat{x}_{T-1}$ are the highest weighted samples from messages received from the $T-1$ other variable nodes connected to $f$.

Each type of constraint node has a unique constraint function for assessing the weight of a sample.  

\textbf{Constraint function example:}
Consider a collision-free constraint node $f_{Cfree}$ sending a message to an arm-trajectory variable node $X_{Traj}$. The collision-free constraint node has edges to both an arm-trajectory variable node and a grasp pose variable node. To compute the \texttt{Constraint-to-variable} message to be sent to the arm-trajectory variable node, $M$ samples are first drawn from the belief of the arm-trajectory variable node. To weight each arm-trajectory sample, the constraint function $\sigma_{Cfree}$ takes as inputs the sample arm-trajectory $x_{traj} ^i$ and the highest weighted grasp pose sample $\hat{x}_{grasp}$.  $\sigma_{Cfree}$ then computes the weight of the sample arm-trajectory as 
\begin{equation}
    w_{traj} ^i =   C_1 \cdot  \epsilon_1(x_{traj} ^i) + C_2 \cdot \frac{1}{\epsilon_2(x_{traj} ^i, \hat{x}_{grasp})}
\end{equation}

\noindent where $w_{traj} ^i$ is the computed weight of sample $x_{traj} ^i$,  $C_1$ and $C_2$ are user-defined constants, $\epsilon_1$ is a routine that computes the cumulative distance along $x_{traj} ^i$ from obstacles in its environment and $\epsilon_2$ is a routine that computes the distance between the end-effector pose after travelling along $x_{traj} ^i$ and the highest weighted grasp pose $\hat{x}_{grasp}$.

The weights for all arm-trajectory samples are computed and normalized, constituting the \texttt{Constraint-to-variable} message sent to the arm-trajectory variable node.

Some other constraint functions used in this work are the 
\begin{itemize}
\item Kinematic constraint function ($\sigma_{kin}$), which weights robot joint configuration samples and grasp poses based on how kinematically feasible a joint configuration of the robot's arm is if it is holding an object with a particular grasp pose.
\item Stable constraint function ($\sigma_{stable}$), which weights placement poses based on how stable they are. i.e. how geometrically stable an object placed on a surface in a specific pose is.
\item Grasp constraint function ($\sigma_{grasp}$), which weights robot configurations, initial object poses and grasp poses based on how well they are jointly feasible and how well they satisfy later target object poses.
\item Motion constraint function ($\sigma_{motion}$), which weights trajectories, initial configuration and final configuration based on how short the trajectory from the initial configuration to the final configuration is.
\item Grasp stability constraint function ($\sigma_{GraspH}$), which weights grasp poses based on how geometrically stable they are.
\item inBasin constraint function, which weights placement poses in the basin based on how stable they are.
\item inSaucepan constraint function, which weights placement poses in the saucepan based on how stable they are. 

\end{itemize}

\section{Implementation} 
\subsection{Action Schemas}\label{sec:schemas}
Actions that make up a plan are described by continuous action parameters, constraints, preconditions and effects. \textbf{\texttt{preconditions}} are the conditions that need to be True before an action can be executed. \textbf{\texttt{effects}} describe the changes in a subset of the state after an action is executed. Continuous action \textbf{\texttt{parameters}} are the continuous values needed by the robot to execute the action in the world. These include object poses, grasps, robot configurations and trajectories.  \textbf{\texttt{constraints}} must hold true for all continuous parameters for the action to be executed successfully in the world. We used Fast Downward \cite{fd} as the symbolic planner for planning action schemas. The types of constraints used in the composition of action schemas in this work are described in Section \ref{sec:cnstfxn}.


Some examples of action schemas \cite{actionschema} used in this work are as follows:

{\scriptsize
\noindent \texttt{ 
\noindent(\textbf{:action} pick[obj] \\
\noindent\textbf{:parameters} ($\phi$, p, g, $\tau$)\\
\noindent\textbf{:constraints} CFree($\tau$), Stable[obj](p), GraspH[obj](g) Kin[obj]($\phi$, p, g)\\
 \textbf{:preconditions} (and (at robot $\phi$) (at obj p) (handempty))\\
 \textbf{:effects}  (and (holding obj) (not (handempty)))\\
)}

\noindent \texttt{ 
\noindent(\textbf{:action} place[obj] \\
\noindent\textbf{:parameters} ($\phi$, p, g, $\tau$)\\
\noindent\textbf{:constraints} CFree($\tau$), Stable[obj](p), GraspH[obj](g) Kin[obj]($\phi$, p, g)\\
 \textbf{:preconditions} (and (holding obj) (at robot $\phi$) (at obj p))\\
 \textbf{:effects}  (and (not holding obj)  (handempty))\\
)}

\noindent \texttt{ 
\noindent(\textbf{:action} wash[obj] \\
\noindent\textbf{:parameters} (p)\\
\noindent\textbf{:constraints} Stable[obj](p), InBasin[obj](p)\\
 \textbf{:preconditions} (at obj p)\\
 \textbf{:effects}  (clean obj)\\
)}

}
\noindent where {\scriptsize\texttt{obj}} represents the target object and {\scriptsize\texttt{$\phi$, p, g, $\tau$}} represent robot joint-configuration, object pose, grasp and joint trajectory respectively.

\begin{figure}%
    \centering
    \subfloat[\centering ]{{\includegraphics[width=.4\linewidth]{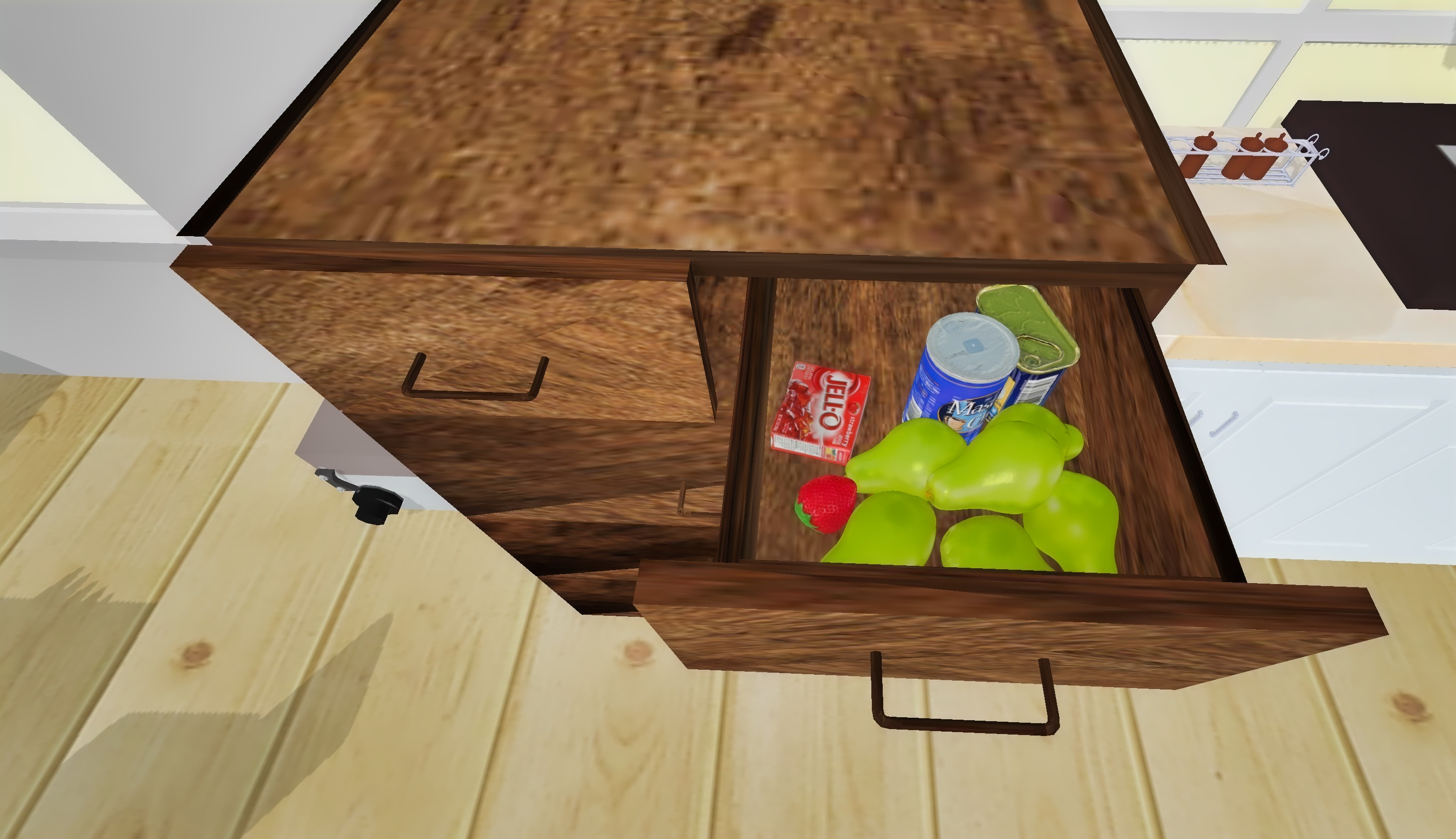} }}%
    \qquad
    \subfloat[\centering ]{{\includegraphics[width=.4\linewidth]{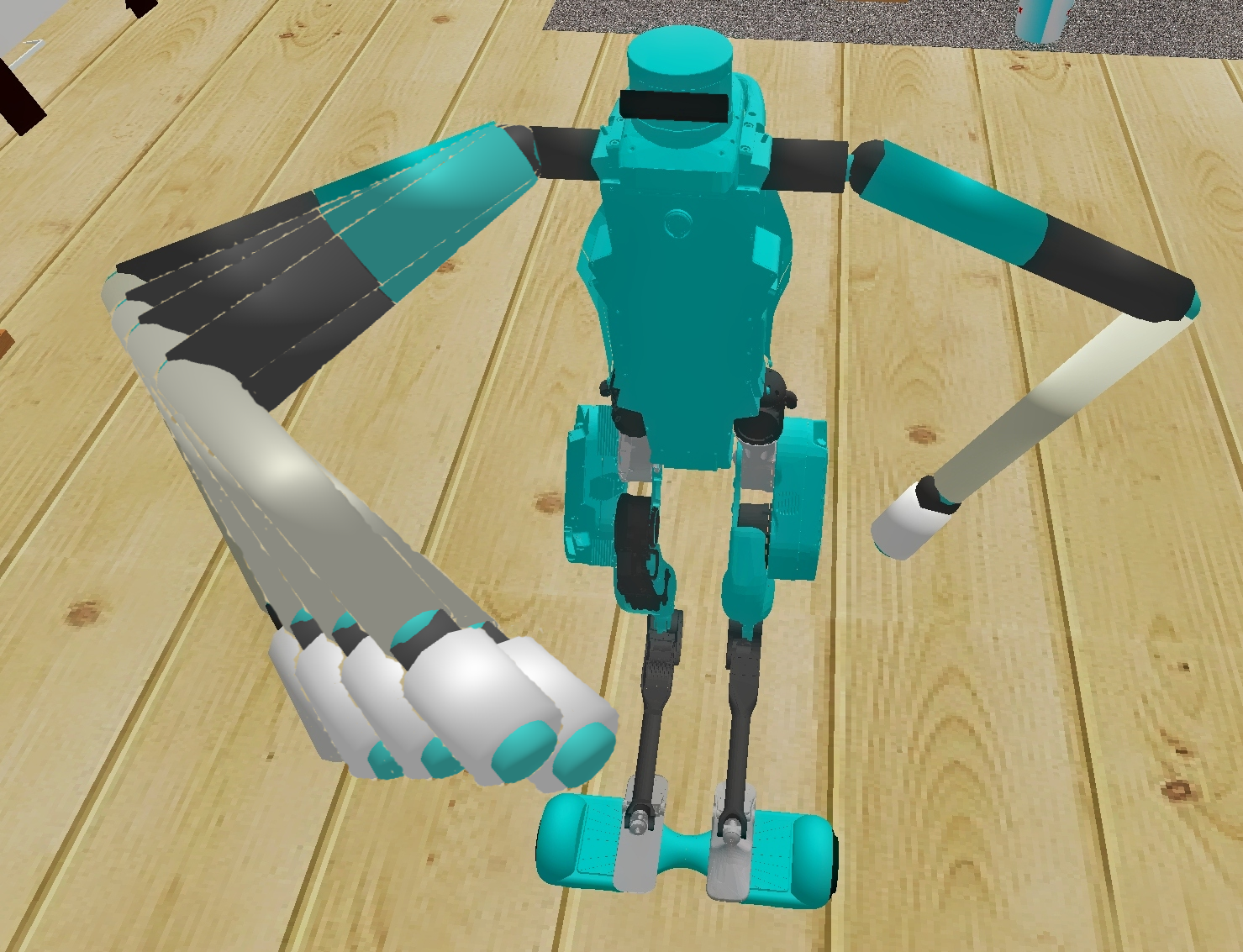} }}%
    \caption{(a) Noisy pose estimate of a single pear in the cabinet drawer. (b) Noisy joint configuration estimate of the robot's right arm. Each pear in (a) and arm configuration in (b) represents a likely pose of the pear or arm joint configuration sampled from their respective estimated noise distributions.}%
    \label{fig:noisiness}%
\end{figure}

\subsection{Uncertainty sources and uncertainty distributions}
The major advantage of \acro{} over Garrett et. al. \cite{beliefpddlstream} is the ability of \acro{} to concurrently and seamlessly incorporate arbitrary uncertainty sources and distributions. In our experiments, we consider pose estimation uncertainty and robot arm joint-configuration uncertainty as depicted in Figure \ref{fig:noisiness}. We assume that the uncertainties in pose estimation and joint configuration estimation are Gaussian distributed with zero means and variances $\delta_p ^2$ and $\delta_c ^2$ respectively. It is worth noting that \acro{} is agnostic to the type of uncertainty distribution of an action parameter variable. 

The robot is equipped with a perception system that updates the robot's belief after every action.

\subsection{Computing initial samples}\label{sec:init_wtsamp} 
We compute initial samples for the various free variables by using their corresponding specialized generators. Each free variable comes with a generator that computes samples for the free variable given the initial belief.

Consider an action that picks up a target object, as described in Section \ref{sec:schemas} above. This action takes as parameters a pose variable, a grasp variable, an arm-trajectory variable and a joint configuration variable.

The initial samples for the pose variable node in the corresponding plan skeleton consist of uniformly weighted poses received from the robot's perception system's noisy estimation of the pose of the object.\\
Likewise, the initial grasp samples are also generated by a grasp generator which computes valid uniformly weighted grasps of the object at each of the initial sample poses. The initial arm-trajectory samples are generated by a specialized arm-trajectory generator equipped with the RRT-Connect \cite{rrtconnect} motion planner. This generator generates uniformly-weighted joint trajectories to each of the grasp samples.

Finally, to generate robot joint configuration samples, the joint-configuration generator, which is equipped with the IKFast \cite{ikfast} inverse kinematics solver, generates joint configuration samples to each of the grasp samples.
With each of these joint configuration samples as a mean, we further sample sub joint configurations from the Gaussian distributed joint configuration noise with variance $\delta_c ^2$. During message passing, the weight $w_i$ of each sample $\phi_i$ is computed as 
\begin{equation*}
    w_i = \sum_{j} p(\phi_i ^j)*\sigma_f(\phi_i ^j, \hat{x}_1, \hat{x}_2, \dots, \hat{x}_{T-1})
\end{equation*}
where $\phi_i ^j$ is the joint configuration sampled from the Gaussian distribution with mean $\phi_i$ and variance $\delta_c ^2$, $p(\phi_i ^j)$ is the Gaussian probability of $\phi_i ^j$, $\sigma_f$ is the constraint function of the constraint node connected to $\phi_i$  and $\hat{x}_1, \hat{x}_2, \dots, \hat{x}_{T-1}$ are the highest weighted samples from messages received from the $T-1$ other variable nodes connected to $f$.
\def\al[#1]{{\color{red}AL:#1}}

\section{Experiments}
We performed experiments on 12 randomly generated  problems for 4 different tasks as described in Section \ref{sec:tasks}. The experiments were performed with a simulated  Digit robot \cite{digit} with the PyBullet simulation software \cite{pybullet} as shown in Figure \ref{fig:serve}. We used IKFast \cite{ikfast} to compute inverse kinematics solutions for the robot arms and used the pybullet\_planning package \cite{pyplan} for motion planning.

We quantitatively compare  \acro{} with \textit{SS-Replan*}, a variation of the SS-Replan algorithm \cite{beliefpddlstream}  which uses off-the-shelf RRT-Connect \cite{rrtconnect} and IKFast \cite{ikfast} for motion planning and inverse kinematics respectively.  

For each cycle of planning, we generate 100 samples from the uncertainty distributions of object pose estimate and 100 samples from the joint configuration estimate.  We run PMPNBP for 10 iterations on each cycle. The pose estimate noise is Gaussian distributed with zero mean and $\delta_p = $ 10cm standard deviation. The joint configuration estimate noise is Gaussian distributed with zero mean and $\delta_c = $ 0.25 radians standard deviation. The experiments were run on a laptop with 2.21GHz Intel Core i7 CPU, 32GB RAM and a GTX 1070 GPU.

\subsection{Tasks}\label{sec:tasks} 
We evaluated  \acro{} and \textit{SS-Replan*} on 12 randomly generated problems of 4 different tasks. 
See the accompanying video for demonstrations of the tasks in simulation. 
The tasks are described as follows:

\subsubsection{Retrieve}
The high-level goal for this task is to retrieve the pear. The prior location of the pear is uniformly distributed across the 3 grocery drawers. Because of occlusion and poor lighting in the drawer, the robot has to deal with the noisy estimates of the pose of the pear as well. A successful plan opens a drawer at random and inspects it. If the pear is located, its done. Else, it updates its belief of the location of the pear and repeats the process until the pear is located. It then picks up the pear.

\subsubsection{Wash}
The high-level goal for this task is to retrieve the pear and wash it. This task has the same prior belief as the task above. A successful plan performs the \textit{Retrieve} task as described above, sends the pear to the sink and turns on the tap to wash it.  

\subsubsection{Cook} 
 The high-level goal for this task is to retrieve the pear, wash it and cook it. The main challenge in this task is to infer grasps and trajectories that allow the robot to pick up a cup in a specific grasp pose that makes later actions like filling the cup with water and pouring the water in the saucepan feasible during the execution of the cooking task. This task has the same prior belief as the tasks above. A successful plan performs the \textit{Wash} task as described above, takes the pear to the stove, puts it in the saucepan, picks up a cup, fills it with water, pours the water in the saucepan and presses the cook button on the stove to cook the pear.

\subsubsection{Serve-meal}
The high-level goal for this task is to retrieve the pear, wash it, cook it and serve it. 
This task has the same prior belief as the tasks above. A successful plan performs the \textit{Cook} task as described above, picks up the cooked pear from the saucepan and sets it on a tray, carries the tray to the dining table and distributes the cooked pear to the plates on the table.

\subsection{Results}\label{sec:results}
Table \ref{table:results} shows the experimental results for  \acro{} and \textit{SS-Replan*} on the tasks described above.
\begin{table}
\centering
\resizebox{\linewidth}{!}{
\begin{tabular}{|l|l|l||l|l|}
\hline 
\textbf{Alg}: & \multicolumn{2}{l|}{\acro} & \multicolumn{2}{l|}{SS-Replan*  }  \\
\hline \hline
\textbf{Task:} & Planning Time & N.E & Planning Time & N.E  \\
  \hline
\textit{Retrieve} & $9.37 \pm 5.25$   & $1.33 \pm 1.52$ & $16.61 \pm 5.68$ & $8.67  \pm 4.93$  \\
\hline
\textit{Wash} & $15.72 \pm 6.91$ & $2.67 \pm 3.06$ & $21.32 \pm 8.53$  & $7.00 \pm 3.60$ \\  
\hline
\textit{Cook} & $25.15  \pm 5.16$   & $3.50  \pm 2.12$   & $29.28  \pm 13.61$   & $6.33 \pm 0.58$ \\
\hline
\textit{Serve-meal} & $37.24 \pm 10.04$ & $7.00  \pm 2.83$ &  $52.75 \pm 16.82$  & $15.5 \pm 0.71$ \\ 
\hline
\end{tabular}}
\caption{Results from evaluation of  \acro{} and SS-Replan*. The table shows the mean planning duration in seconds (Planning Time) and the number of errors (N.E) for all 12 randomly generated problems for the 4 tasks}
\label{table:results}
\end{table}

\begin{figure}
\centering
     \includegraphics[width=0.5\textwidth]{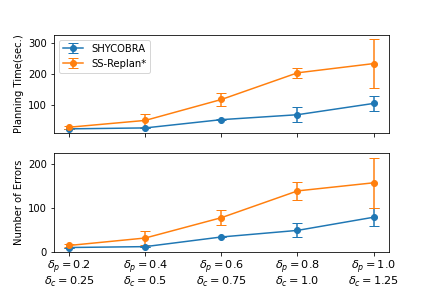}
      \caption{Plots comparing the planning times and number of errors of \acro{} and SS-Replan* for increasing uncertainties in the pose estimation and joint configuration when performing the \textit{Retrieve} task. $\delta_p$ represents the standard deviation of the zero mean object pose estimation noise  and $\delta_c$ represents the standard deviation of the zero mean joint configuration estimation noise }
      \label{fig:alg_uncertainty}
\end{figure}

An error occurs when the robot misses its target when picking or placing an object due to noise in the object's pose or joint configuration estimate. 
Based on the results, \acro{} was consistently more efficient than \textit{SS-Replan*} across all 4 tasks and made slightly less errors across all 4 tasks.  Since \textit{SS-Replan*} is only capable of considering object pose uncertainty and doesn't account for the uncertainty in the robot arm's joint configurations, it misses its target more often and as a result makes more errors and takes longer to plan.

In spite of its ability to concurrently incorporate arbitrary uncertainty sources and distributions due to the use of PMPNBP for inference, \acro{} has a time complexity that  grows with the magnitude of noise in the estimation. We demonstrate this by comparing the planning times for SHY-COBRA and SS-Replan* for increasing noise in object pose estimation and joint configuration estimation as indicated in the results in Figure \ref{fig:alg_uncertainty}.  
\section{Conclusion} 

We proposed a planning approach for long horizon planning under uncertainty. Our approach jointly infers satisfying action parameter values for a plan skeleton that are needed to successfully execute the plan in a stochastic, partially observable environment. Our approach outperformed an adaption of the SS-Replan algorithm across all tasks in our experiments.
 
\bibliographystyle{plain}
\bibliography{references}

\end{document}